# A Comprehensive Part-of-Speech Tagging to Standardize Central-Kurdish Language: A Research Guide for Kurdish Natural Language Processing Tasks

*Shadan Shukr Sabr[1], Nazira Sabr Mustafa[1], Talar Sabah Omar[1], Salah Hwayyiz Rasool [1], Nawzad Anwer Omer[1], Darya Sabir Hamad[1], Hemin Abdulhameed Shams[1], Omer Mahmood Kareem [1], Rozhan Noori Abdullah[1], Khabat Atar Abdullah[2], Mahabad Azad Mohammad[1], Haneen Al-Raghefy[3], Safar M. Asaad[3,4], Sara Jamal Mohammed[5], Twana Saeed Ali[6], Fazil Shawrow[1], and Halgurd S. Maghdid[7]*

*[1]Department of Kurdish Language, Faculty of Education, Koya University, Danielle Mitterrand Boulevard, Koya, KOY45, Kurdistan Region – F.R. Iraq.*
*[2]Department of Computer Science, Faculty of Health and Science, Koya University, Koya KOY45, Kurdistan Region – F.R. Iraq*
*[3]Department of Software Engineering, Faculty of Engineering, Koya University, Danielle Mitterrand Boulevard, Koya, KOY45, Kurdistan Region – F.R. Iraq.*
*[4]Department of Computer Engineering, College of Engineering, Knowledge University, Erbil 44001, Iraq.*
*[5]Department of English language, Faculty of Education, Koya University, Danielle Mitterrand Boulevard, Koya, KOY45, Kurdistan Region – F.R. Iraq.*
*[6]Department of Computer Engineering, College of Engineering, University of Sulaimani, Kurdistan, Region, Iraq.*
*[7]Department of Engineering Research Center, Deanship of R&D Centers, Koya University, Danielle Mitterrand Boulevard, Koya, KOY45, Kurdistan Region – F.R. Iraq.*

Author: Shadan Shukr Sabr from Department of Kurdish. Faculty of Education at Koya University, Kurdistan Region, Iraq.
Email: shadan.shukr@koyauniversity.org
ORCID: https://orcid.org/0000-0002-4444-934X
GSC: https://scholar.google.com/citations?hl=en&user=8wJ-uUAAAAAJ&view_op=list_works&gmla=AJsN-F4UaK-i-S3ruYxuwVz3_CNzM0nGV_CT3ezWfZbTCxq6h_bITAN90yQ0UO
Author: Nazira Sabr Mustafa from Department of Kurdish Language, Faculty of Education, Koya University.
Email: nazira.sabr@koyauniversity.org
ORCID: https://orcid.org/0000-0002-0201-9026
GSC: https://scholar.google.com/citations?hl=en&user=QdhgNwwAAAAJ&view_op=list_works&sortby=title
Author: Talar Sabah Omar from Department of Kurdish Language, Faculty of Education, Koya University, Koya Koy45, Iraq.
Email: talar.sabah@koyauniversity.org
ORCID: https://orcid.org/0000-0002-6247-4995
GSC: https://scholar.google.com/citations?user=DZD9LB8AAAAJ&hl=en
Author: Salah Hwayyiz Rasool from Department of Kurdish Language, Faculty of Education, Koya University, Koya Koy45,Iraq.
Email: salah.hawez@koyauniversity.org
ORCID: https://orcid.org/0000-0002-8556-8644
GSC: https://scholar.google.com/citations?user=aNcGGm0AAAAJ&hl=en
Author: Nawzad Anwer Omer from Department of Kurdish Language, Faculty of Education, Koya University, Koya KOY45, Iraq.
Email: nawzad.anwer@koyauniversity.org
ORCID: https://orcid.org/0000-0002-4188-7635
GSC: https://scholar.google.com/citations?user=VYHCKdwAAAAJ&hl=en
Author: Darya Sabir Hamad from Department of Kurdish Language, Faculty of Education, Koya University, Koya KOY45, Iraq.
Email: darya.sabr@koyauniversity.org
ORCID: https://orcid.org/0000-0002-4226-7016
GSC: https://scholar.google.com/citations?user=flqq0PEAAAAJ&hl=en
Author: Hemin Abdulhameed Shams from Department of Kurdish Language, Faculty of Education, Koya University, Koya KOY45, Iraq.
Email: hemin.shams@koyauniversity.org

ORCID: https://ORCID.org/0000-0003-3866-581X
GSC: https://scholar.google.com/citations?hl=en&user=wcXwEOYAAAAJ
Author: Omer Mahmood Kareem from Department of Kurdish Language, Faculty of Education, Koya University, Koya KOY45, Iraq.
Email: omar.mahmood@koyauniversity.org
ORCID: https://orcid.org/0000-0002-6258-342X
GSC: https://scholar.google.com/citations?user=bF7dfxYAAAAJ&hl=en
Author: Rozhan Noori Abdullah from Department of Kurdish Language, Faculty of Education, Koya University, Koya KOY45, Iraq.
Email: rozhan.noori@koyauniversity.org
ORCID: https://orcid.org/0000-0002-5219-0235
GSC: https://scholar.google.com/citations?view_op=list_works&hl=en&user=BJrOhK4AAAAJ
Author: Khabat Atar Abdullah from Department of Computer Science, Faculty of Health and Science, Koya University, Koya KOY45, Kurdistan Region – F.R. Iraq.
email: khabat.atar@koyauniversity.org
ORCID link: https://orcid.org/0009-0005-7008-9041
GSC: https://scholar.google.com/citations?user=Ytt_Z4YAAAAJ&hl=en
Author: Mahabad Azad Mohammad from Department of Kurdish Language, Faculty of Education, Koya University, Koya Koy45, Kurdistan Region – F.R. Iraq.
Email: mahabad.azad@kouauniversity.org
ORCID: https://orcid.org/0000-0002-1170-7795
GSC: https://scholar.google.com/citations?view_op=list_works&hl=en&hl=en&user=-8v3laYAAAAJ
Author: Haneen Al-Raghefy from Department of Software Engineering, Faculty of Engineering, Koya University, Danielle Mitterrand Boulevard, Koya, KOY45, Kurdistan Region – F.R. Iraq.
Email: haneen.hayder@koyauniversity.org
ORCID: https://orcid.org/0009-0007-1867-6614
GSC: XXX
Author: Safar M. Asaad from Department of Software Engineering, Faculty of Engineering, Koya University, Danielle Mitterrand Boulevard, Koya, KOY45, Kurdistan Region – F.R. Iraq. And he is also from Department of Computer Engineering, College of Engineering, Knowledge University, Erbil 44001, Iraq.
Email: safar.maghdid@koyauniversity.org
ORCID: https://orcid.org/0000-0003-0808-7545
GSC: https://scholar.google.com/citations?hl=en&user=LhrjERAAAAAJ
Author: Sara Jamal Mohammed from Department of English language, Faculty of Education, Koya University, Danielle Mitterrand Boulevard, Koya, KOY45, Kurdistan Region – F.R. Iraq
Email: sara.jamal@koyauniversity.org
ORCID: https://orcid.org/my-orcid?orcid=0000-0002-8510-7805
GSC: https://scholar.google.com/citations?user=Wy3q4CcAAAAJ&hl=en
Author: Twana Saeed Ali from Department of Computer Engineering, College of Engineering, University of Sulaimani, Kurdistan, Region, Iraq.
Email: twana.ali@univsul.edu.iq
ORCID: https://orcid.org/0000-0003-4931-9939
GSC: https://scholar.google.com/citations?user=arQZhsIAAAAJ&hl=en&oi=ao
Author: Fazil Shawrow from Department of Kurdish Language, Faculty of Education, Koya University, Koya KOY45, Iraq.
Email: fzgaedi@gmail.com
Author: Halgurd S. Maghdid from Department of Engineering Research Center, Deanship of R&D Centers, Koya University, Danielle Mitterrand Boulevard, Koya, KOY45, Kurdistan Region – F.R. Iraq
Email: halgurd.maghdid@koyauniversity.org
ORCID: https://orcid.org/ 0000-0003-1109-4009
GSC: https://scholar.google.com/citations?hl=en&authuser=1&user=yjtMyygAAAAJ

***Abstract*** - The field of natural language processing (NLP) has dramatically expanded within the last decade. Many human-being applications are conducted daily via NLP tasks, starting from machine translation, speech recognition, text generation and recommendations, Part-of-Speech tagging (POS), and Named-Entity Recognition (NER). However, low-resourced languages, such as the Central-Kurdish language (CKL), mainly remain unexamined due to shortage of necessary resources to support their development. The POS tagging task is the base of other NLP tasks; for example, the POS tag set has been used to standardized languages to provide the relationship between words among the sentences, followed by machine translation and text recommendation. Specifically, for the CKL, most of the utilized or provided POS tagsets are neither standardized nor comprehensive. To this end, this study presented an accurate and comprehensive POS tagset for the CKL to provide better performance of the Kurdish NLP tasks. The article also collected most of the POS tags from different studies as well as from Kurdish linguistic experts to standardized part-of-speech tags. The proposed POS tagset is designed to annotate a large CKL corpus and support Kurdish NLP tasks. The initial investigations of this study via comparison with the Universal Dependencies framework for standard languages, show that the proposed POS tagset can streamline or correct sentences more accurately for Kurdish NLP tasks.

## 1- Introduction

Language is one of the essential factors in forming communication activity and expressing human thoughts and perspectives about the surrounding world. Language is the most crucial characteristic that distinguishes humans from other living beings. Every day, human beings use language to communicate with one another. Language is also interconnected with various fields of science, which is why it has many different definitions. For example, ***Ferdinand de Saussure*** **says**, "Language is a social product of the speech community and a collection of necessary conventions, adopted by society, to help individuals communicate" [1]**.** Similarly**,** ***Khuli*** states that "Language is a system of relationships between two layers" [2]**.** That is to say, language is an independent symbolic system that forms within society and is used as a tool for mutual understanding among individuals. Additionally, language is essential in shaping and transmitting the culture and values of society, preserving these elements over time. Language, thought, and human actions are deeply interconnected in social life, forming an inseparable bond [3]**.** Language is a tool for mutual understanding and the exchange of thoughts. In every layer of society and every interaction, we see language present, alive, and essential, becoming such an integral part of human existence that life without it is unimaginable. Civilization itself cannot continue without language, as it is the only means of preserving and transmitting culture and knowledge from one generation to another [4].

In another vein, within the 21$^{st}$ century, most of the daily human beings' activities, including the aforementioned activities, are run via natural language processing (NLP) tasks. The range of NLP tasks starts from machine translation, text translation, spell checking, similarities, speech recognition, POS tagging, and text generation [5]. In particular, POS tagging is the base of other NLP tasks, such as spell-checking and text generation [6], [7]. The worldwide spoken languages, including English, Spanish, Persian, and others, are fully utilized via NLP tasks. This is due to the existing comprehensive annotated corpora and the development of large language models (LLM) [8], [9]. Furthermore, with the era of today's generative artificial intelligence (AI), the existing AI algorithms and huge cloud data/documents have essential role to train transformers or LLMs to generate accurately text, voices, and image without using POS tagging. However, for Kurdish Language within all dialects, the developed LLM models are not well-trained yet. This is because, 1) the Kurdish text is not massively available in the cloud, and 2) even the available Kurdish text or data is not accurate due to not existing a standardized morphological/syntactic.

Therefore, the Kurdish community and Kurdish NLP tasks need an accurate POS tagset to annotate massive or big corpora. To this end, several researchers and commercial companies are engaged in creating and developing such big corpora and followed by training the data to construct accurate and applicable LLM models. However, we can have noticed that their models are not well utilized via Kurdish NLP tasks. This is because of not using an accurate POS tagset and not existing a big corpus. For example, although frameworks like Universal Dependencies (UD) [10], [11] offer a general framework for POS tags, however, the UD doesn't include the level of detail necessary for the annotation of some language-specific phenomena such as for Kurdish language. On the other hand, in terms of both morphological and syntactical perspectives, the Central Kurdish Language (CKL) is a linguistically rich language. To capture all these issues, there should be an accurate POS tagset to annotate big corpus, and then utilizing the annotated corpus to formulate the morphological/syntactic rules Kurdish text. Thereafter, the existing LLM models can benefit from this process to run the Kurdish NLP task.

Therefore, this study presents a research guideline to provide an accurate POS tagset for CKL. This is followed by constructing the POS tagset based on existing frequently utilized POS tags or labels the CKL words/tokens in the CKL text. To the best of our experts' knowledge, the proposed POS tagset is a new and comprehensive tagset which enables the Kurdish NLP tasks to be more accurate. The rest of the article is structured as follow, section 2 introduce the richness Central-Kurdish Language (CKL) in terms of diversity of the dialects and speech style. Section 3 investigates current studies of POS tagging for CKL language and lists a set of POS tagging challenges. While, the section 4 presents the proposed POS tagset for CKL as well as explain how the CKL text is analyzed according to CKL word components and part of speech. Finally, the section 5 discusses how the proposed POS tagset would be utilized for NLP tasks and recommends that how the proposed POS tagset can be used for annotating CKL corpus.

## 2- Overview of CKL

Every society has its own distinct language and dialect. Each nation's language is divided into multiple dialects and varieties of speech styles. This is especially true for the Kurdish language, which has branched into several dialects and sub-varieties. Each of these has its unique characteristics and differs from the others. This means that a dialect refers to a variant of a language spoken in a specific geographic region, and a language can consist of multiple dialects. In Kurdish, in addition to dialects, other elements such as accents, speech patterns, and intonation are also included under the term “dialect” [12]**.** The concept of dialect refers to the differences that exist between the variations of the same language. While a language itself remains stable and unchanged, every language in the world has its own distinct dialects and speech forms. Moreover, when a language is spoken by a community within the same linguistic framework, any noticeable variations within that framework are considered dialects.

### *A. The Dialects*

The concept of dialect has been defined in several ways. Here, we briefly highlight a few definitions:

- A dialect is a linguistic characteristic tied to a specific environment, where all individuals in that environment participate in those characteristics. It is a form of language used by a defined group within a community that possesses its own vocabulary, grammar, and unique features distinct from those of other groups within the same community.
- In other words, a dialect refers to the language used by a group of people in a specific location, and it is often marked by their specific expressions, even though they may be able to understand the standard language. Furthermore, a dialect is a localized or regionally specific form of language, often used by a distinct group within a society that speaks a broader, more general language. These dialects can vary depending on the region and community, reflecting local cultural and social differences [12].
- A dialect may also be referred to as the variety of language used in a particular geographic area or by a specific social group within the larger community.

From these definitions, it is also understood that a dialect refers to the different ways of speaking within the boundaries of one language, which serves as an identity marker for the speaker. A language covers a wide geographical area, while a dialect is specific to a defined region. This means that the entirety of a language encompasses the differences found in its dialects.

Like any other language in the world, the Kurdish language comprises several dialects [2]. The relationship between language and dialect is general, meaning that a dialect is a part of a language and consists of multiple dialects. As a living language, Kurdish consists of many dialects and speech styles. Furthermore, the Kurdish language can be divided into four main dialects, and these primary dialects are further divided into regional dialects, as shown in figure 1. These divisions are as follows [13]:

1. ***Northern Kurmanji Dialect:*** Including regions such as ***Bayazid, Hakari, Botani, Şemdinan, Behdinan***, and the dialects of the West.
2. ***Central Kurmanji Dialect*:** Includes ***Mukri, Sorani, Erdelani, Sulaymani*** and ***Garmiani***, with study, we called Central Kurdish Language (CKL).
3. ***Southern Kurmanji Dialect*:** including ***Luri Rasan, Bakhtiari, Mamassani, Koglu, Lak*** and ***Kalhor***.
4. ***Goran Dialects:*** Include Goran ***Rasan, Hawramani, Bajelani,*** and ***Zaza*** *[14]***.**

Considering our study specifically on Central Kurmanji Dialect, we have to further divide this dialect into more specific sub-dialects and note the regions where the speakers reside:

- ***Mukri sub-dialect:*** spoken in areas such as ***Sheno, Naghdah, Maragheh, Miandav, Shahin Dej, Saqqez, Bokan, Baneh, Sardasht*** and ***Mahabad.***
- ***Sorani sub-dialect:*** spoken in the ***Zarrabari*** region and all of Erbil province, including the city of Erbil.
- ***Erdelani sub-dialect:*** spoken in the areas of ***Bijar, Kangawar, Ravansar,*** and the northern parts of the ***Jwanro*** area, including the ***Sanandaj city***.
- ***Sulaymaniyah sub-dialect*:** Spoken throughout ***Sulaymaniyah*** province, as well as parts of the ***Khanaqin*** district.
- ***Garmiani sub-dialect*:** Spoken in regions like ***Kifri***, ***Qaradagh***, ***Kirkuk***, ***Goz*** and ***Shwani*** [14]**.**

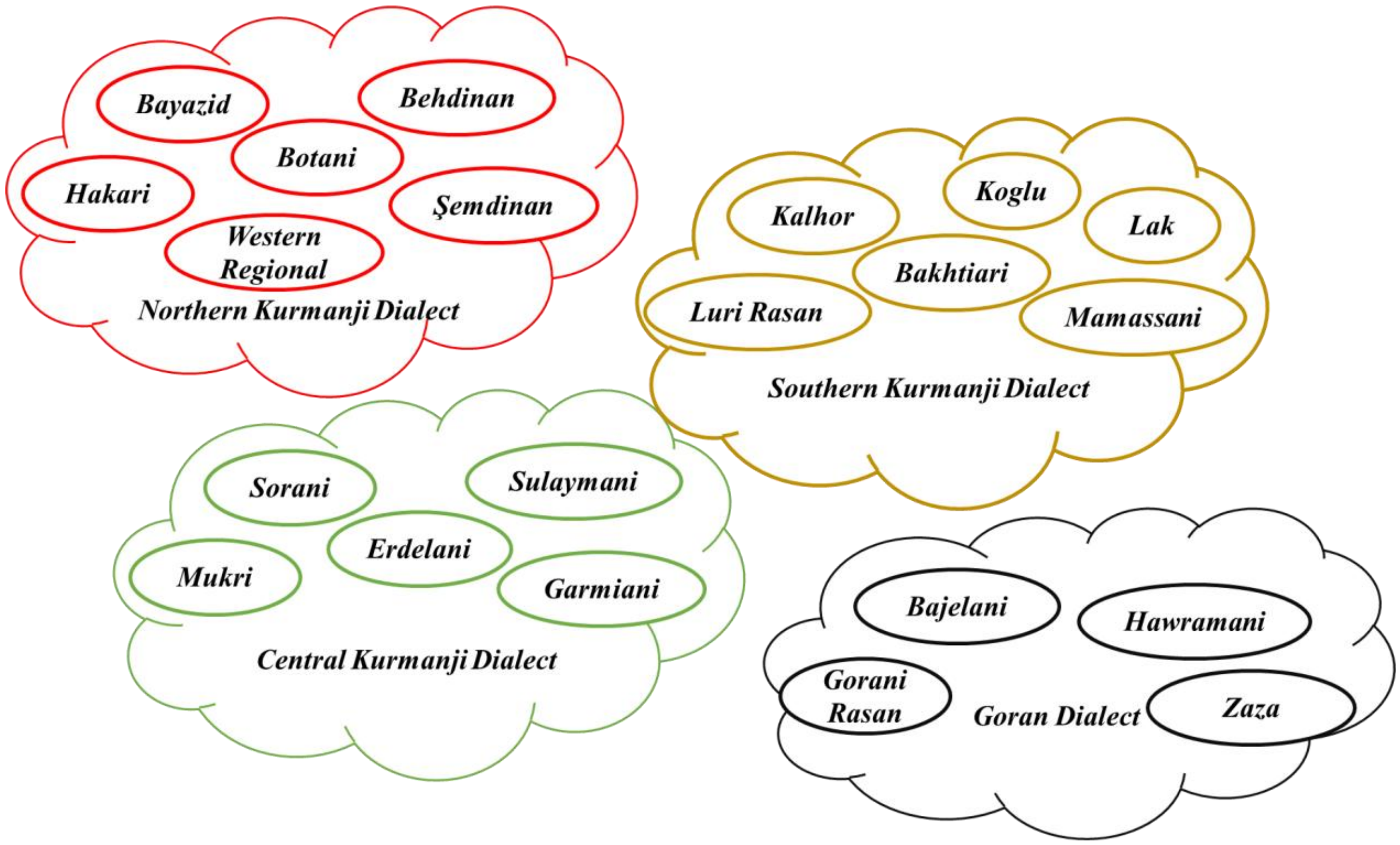

*Figure 1. The distributed sub-dialects according to Kurdish Language dialects.*

The other characteristic of language is the speech style. The speech style refers to people's distinct vocal and lexical variations within a specific geographic and linguistic area. Linguists have referred to these variations using different terms depending on the language. For example, in Kurdish, terms like ***"Sorani sub-dialect"***,"***Slemany sub-dialect***", ***"speech style"***, and "***dialect***" are used. In Arabic, it is referred to as "**اللكنة**" (***pronunciation***), in Persian as "**لهجه**" (***accent***), and in English, the term "***accent***" is commonly used *[3]***.**

### *B. The Speech Style*

The speech style encompasses all the pronunciation and word usage differences that people in a particular region exhibit when speaking. These differences might show similarities with neighboring areas but can also be unique to specific regions. This variation in speech is an essential characteristic for identifying and distinguishing people from different regions.

Based on this concept, every city, town, or village has its own specific speech pattern [15]. It is also noted that speech style refers to the gradual changes in the characteristics of speech that help define an individual's identity based on their community and geographic origin. Linguists emphasize that speech style is primarily concerned with pronunciation and phonetic variation [16]. In addition, the linguists noticed two important aspects regarding speech style in this context. The **first** point is that it is only related to pronunciation, and the **second** is that it is universal, meaning every individual speaks in a particular style, which includes the way they pronounce sounds and use words within a specific context [17]. For example, in Kurdish, the pronunciation of the sounds (ڵ) and (ر) in the word "***ماڵ***" (**home**) is one of the distinguishing features between different speech styles, such as the difference between **Kurmanji dialects**. In some dialects, (**مار**) or the sound (ر) in the same word may be pronounced as (**مار**), showing a variation within the same word. This illustrates that when a speaker uses a speech style, they reflect the characteristics of their dialect, which is scientifically understood as a variant of the "speech style" representing how the speaker identifies with their region and community [18]**.** In comparison, the same issue of speech style exists in the United Kingdom. For instance, in the region known as "***the West***", the "***Liverpool***" accent is a defining feature. Additionally, speakers of the same language, such as English, can exhibit different speech styles depending on the region, as seen in the differences between Australian and American accents.

It is important to note that there is a clear distinction between dialect and speech style. Speech style refers to variations in phonetic and phonological features, while dialect involves syntax and word formation differences. Unlike dialect, which often includes syntactic and lexical differences, speech style focuses on phonetic and acoustic features, reflecting the speakers' geographic area and social characteristics. These variations in speech style highlight the social context in which the language is used [15]**.** Linguists also emphasize that speech style is primarily connected to pronunciation, which distinguishes it from dialect. Dialects, on the other hand, involve differences not only in pronunciation but also in grammar and vocabulary. This distinction is important because speech style tends to reflect regional or social identity more directly, whereas dialects encompass broader linguistic features [16]**.**

The aforementioned characteristics and the richness of the ***Central Kurmanji Dialect*** (or the Central Kurdish Language) provide a new challenge to standardizing the syntactic rules. This is followed by constructing a comprehensive and accurate POS tagset. This is because the syntactic rules and labels/tags of the words within the sentences should be defined within the NLP tasks. To tackle this issue, this study presents a new accurate POS tagset and applies it to a big CKL corpus.

## 3- Related Work

The Kurdish NLP tasks need a very fine-grained and deep part-of-speech (POS) tagging. This is due to the fact that the CKL is a morphologically and syntactically rich language. Although frameworks like UD [10], [11] offer a general framework, they do not include the level of detail necessary for the annotation of some language-specific phenomena in Kurdish.

Previous studies on Kurdish POS tagging have also addressed the POS tagging of Kurdish and defined local linguistic categories. The studies of Awrahmani Haji Marf have focused on some other aspects of Kurdish morphology and syntax, such as nouns [22], pronouns [23], and verbs [24]. Although these studies are essential for politeness in Kurdish, they fail to provide a structured POS tagset that can be used for NLP. In contrast, we present a consistently fine-grained Kurdish POS tagset that we developed in consultation with experts in the Kurdish language [25]. This approach makes classifying morphosyntactic properties more detailed, systematic, modular, and compatible with modern NLP frameworks while maintaining the distinctive nature of Kurdish linguistics.

In another vein, with a recent study, a Kurdish POS tagset of 38 tags could not cover the richness of CKL morphology and syntax [26]. For instance, it omitted distinctions between material and abstract nouns or verbs of various sorts (e.g., causative, passive). To counter this issue, we have drastically increased the tagset to 97 tags. With this study, finer distinctions like proper nouns, material nouns, abstract nouns, causative verbs, passive verbs, and comparative adjectives are introduced to facilitate linguistic analysis for CKL language processing using NLP tasks. Thus, by considering these explanations, the new tagset achieves a better consistency with the improved structures in modern NLP while keeping up the CKL's very own linguistic structures.

To address this gap, we propose a fine-grained Central Kurdish POS tagset of 97 tags capturing the morph-syntactic properties of CKL. An advantage of our tagset is that it provides a finer classification for nouns (proper, material, abstract, etc.), it distinguishes different types of verbs (transitive, causative, passive, etc.) as well as more refined types of adjectives (qualitative, superlative, comparative, etc.). This enables us to preserve some linguistic properties of CKL while staying up to date with recent NLP tasks.

## 4- Proposed CKL POS Tagset

According to the previous studies, we investigated that the studies have presented different CKL POS tagset which are neither comprehensive nor standardized. Furthermore, the current state of the art needs further research to standardize CKL POS tagging. We also noted that the previous studies have more focused on multi-dialects with non-comprehensive POS tagging. To this end, this study proposed new and finer distinctions among the POS tags via analyzing morphological aspects, which is to identify the parts of speech (nouns, pronouns, pronouns, prefixes, verbs, numbers, verbs, particles) in the context of syntax. This is because one of the characteristics of CKL language is that a single word can fall into three different word categories, but is separated within a sentence. This is aimed to facilitate the linguistic analysis for CKL language processing using NLP tasks. Our proposed CKL POS tagset is also organized in main linguistic categories with subclasses that capture syntactic and morphological variations, as shown in Table I. Note, for the sake of using the POS tagging within the Kurdish NLP tasks, each tag name is abbreviated, accordingly.

*TABLE I. Categorized the proposed POS Tagset within CKL and equivalent the English meaning.*

| NO | Category | Tag Name (English) | Abbreviation | Tag Name (Kurdish) | NO | Category | Tag Name (English) | Abbreviation | Tag Name (Kurdish) |
|---|---|---|---|---|---|---|---|---|---|
| 1 | Noun | Simple Noun | N-S | ناوی سادە | 50 | Adjective | Adjective of degree | ADJ-DEG | ئاوەڵناوی رادەی |
| 2 | Noun | Compound Noun | N-COMP | ناوی لێکدراو | 51 | Adjective | Relative Adjective | ADJ-REL | ئاوەڵناوی ڕێژەی |
| 3 | Noun | Extended Noun | N-EXT | ناوی داڕێژراو | 52 | Adjective | Indefinite Adjective | ADJ-INDEF | ئاوەڵناوی نادیار |
| 4 | Noun | Proper Noun | N-PROP | ناوی تایبەتی | 53 | Adjective | Descriptive Adjective | ADJ-DESC | ئاوەڵنای چۆنیەتی |
| 5 | Noun | Common Noun | N-COM | ناوی گشتی | 54 | Adjective | Quantitative Adjective | ADJ-QUANT | ئاوەڵناوی چەندیەتی |
| 6 | Noun | Collective Noun | N-COLL | ناوی کۆمەڵ | 55 | Adjective | Fixed Level Adjective | ADJ-BASE | ئاوەڵناوی چەسپاو |
| 7 | Noun | Singular Noun | N-SG | ناوی تاک | 56 | Adjective | Comparative | ADJ-COMP | ئاوەڵناوی پلەی بەراورد |
| 8 | Noun | Plural Noun | N-PL | ناوی کۆ | 57 | Adjective | Superlative | ADJ-SUPL | ئاوەڵناوی پلەی باڵا |
| 9 | Noun | Masculine Noun | N-M | ناوی نێر | 58 | Adverb | Simple Adverb | ADV-S | ئاوەڵکاری سادە |
| 10 | Noun | Feminine Noun | N-F | ناوی مێ | 59 | Adverb | Compound Adverb | ADV-COMP | ئاوەڵکاری لێکدراو |
| 11 | Noun | Dual-Gender Noun | N-DUAL | ناوی دولایەن | 60 | Adverb | Extended Adverb | ADV-EXT | ئاوەڵکاری داڕێژراو |
| 12 | Noun | Gender-Neutral Noun | N-GN | ناوی بێلایەن | 61 | Adverb | Conjunction Adverb | ADV-CONJ | ئاوەڵکاری لێکچواندن |
| 13 | Noun | Concrete Noun | N-CN | ناوی مادی | 62 | Adverb | Manner Adverb | ADV-MAN | ئاوەڵکاری چۆنیەتی |
| 14 | Noun | Abstract Noun | N-AB | ناوی واتایی | 63 | Adverb | Quantitative Adverb | ADV-QUANT | ئاوەڵکاری چەندیەتی |
| 15 | Noun | Definite Noun Phrase | N-DNP | ناوی ناسراو | 64 | Adverb | Negation Adverb | ADV-NEG | ئاوەڵکاری نەفی |
| 16 | Noun | Indefinite Noun Phrase | N-INP | ناوی نەناسراو | 65 | Adverb | Emphatic Adverb | ADV-EMPH | ئاوەڵکاری جەختکردنەوە |
| 17 | Noun | Subject Noun | N-SUB | ناوی بکەر | 66 | Adverb | Adverb of Frequency | ADV-REP | ئاوەڵکاری دوبارەکرنەوە |
| 18 | Pronoun | Independent Pronoun | PR-INDEP | جێناوی سەربەخۆ | 67 | Adverb | Purposeful Adverb | ADV-CAUS | ئاوەڵکاری هۆ و مەبەست |
| 19 | Pronoun | Clitic Pronoun | PR-CLIT | جێناوی لکاو | 68 | Adverb | Adverb of Place | ADV-LOC | ئاوەڵکاری شوێن |
| 20 | Pronoun | Demonstrative Pronoun | PR-DEM | جێناوی نیشانە | 69 | Adverb | Adverb of Time | ADV-TIME | ئاوەڵکاری کاتی |
| 21 | Pronoun | Reflexive Pronoun | PR- REF | جێناوی خۆیی | 70 | Numeral | Simple Number | NUM-S | ژمارەی سادە |
| 22 | Pronoun | Possessive Pronoun | PR-POSS | جێناوی هەیی | 71 | Numeral | Compound Number | NUM-COMP | ژمارەی لێکدراو |
| 23 | Pronoun | Interrogative Pronoun | PR-INT | جێناوی پرسیار | 72 | Numeral | Extended Number | NUM-EXT | ژمارەی داڕێژراو |
| 24 | Pronoun | Negative Pronoun | PR-NEG | جێناوی نەرێ | 73 | Numeral | Cardinal | NUM-CARD | ژمارەی بنجی |
| 25 | Pronoun | Quantifier Pronouns | PR-QUAN | جێناوی چەندێتی | 74 | Numeral | Fractional | NUM-FRAC | ژمارەی کەرتی |
| 26 | Pronoun | Definite Pronoun | PR-DEF | جێناوی دیار | 75 | Numeral | Ordinal | NUM-ORD | ژمارەی ڕێکخستن |
| 27 | Pronoun | Indefinite Pronoun | PR-INDEF | جێناوی نادیار | 76 | Verb | Simple Verb | V-S | کاری سادە |
| 28 | Pronoun | Simple Pronoun | PR-S | جێناوی سادە | 77 | Verb | Extended Verb | V-EXT | کاری داڕێژراو |
| 29 | Pronoun | Compound Pronoun | PR-COMP | جێناوی لێکدراو | 78 | Verb | Compound Verb | V-COMPOUND | کاری لێکدراو |
| 30 | Pronoun | Extended Pronoun | PR-EXT | جێناوی داڕێژراو | 79 | Verb | Past Tense | V-PAST | ڕابردو |
| 31 | Particles | Topical Particle | PART-TOP | ئامرازەکانی خستنەسەر | 80 | Verb | Non-Past Tense | V-NONPAST | ڕانەبردو |
| 32 | Particles | Adverbial Particle | PART-ADV | ئامرازی ئاوەڵکاری | 81 | Verb | Perfective | V-PERF | تێپەڕ |
| 33 | Particles | Locative Particle | PART-LOC | ئامرازە لێکترازاوەکان | 82 | Verb | Imperfective | V-IMPERF | تێنەپەڕ |
| 34 | Particle | Conjunction Particle | PART-CONJ | ئامرازەکانی بەستنەوە | 83 | Verb | Complete | V-COMPLETE | تەواو |
| 35 | Particles | Interrogative Particle | PART-INT | ئامرازی پرسیار | 84 | Verb | Incomplete | V-INCOMPLETE | ناتەواو |
| 36 | Particles | Conditional Particle | PART-COND | ئامرازی مەرج | 85 | Verb | Affirmative | V-AFF | ئەرێ |
| 37 | Particles | Request Particle | PART-REQ | ئامرازی داخوازی | 86 | Verb | Negative | V-NEG2 | نەرێ |
| 38 | Particles | Surprise Particle | PART-SURP | ئامرازی سەرسوڕمان | 87 | Verb | Informative | V-INFORM | ڕاگەیاندن |
| 39 | Particles | Exclamation Particle | PART-EXCL | ئامرازی بانگکردن | 88 | Verb | Declarative | V-DECL | دانانی |
| 40 | Particles | Softeners & Politeness | PART-POL | ئامرازی توانج و سوکایەتیپێکردن | 89 | Verb | Imperative | V-IMPER | فەرماندان |
| 41 | Particles | Emphatic Particle | PART-EMPH | ئامرازی جەختکردنەوە | 90 | Gerund | Gerund Dalî | GRD-D | چاوگی دالی |
| 42 | Adjective | Simple Adjective | ADJ-SIMPLE | ئاوەڵناوی سادە | 91 | Gerund | Gerund Wawî | GRD-W | چاوگی واوی |
| 43 | Adjective | Extended Adjective | ADJ-EXT | ئاوەڵناوی دارێژراو | 92 | Gerund | Gerund Yayî | GRD-Y | چاوگی یائی |
| 44 | Adjective | Compound Adjective | ADJ-COMPOUND | ئاوەڵناوی لێکدراو | 93 | Gerund | Gerund Tayî | GRD-T | چاوگی تائی |
| 45 | Adjective | Dimensional Adjective | ADJ-DIM | ئاوەڵناوی بارستایی(قەبارە) | 94 | Gerund | Gerund Alfî | GRD-A | چاوگی ئەلفی |
| 46 | Adjective | Color Adjective | ADJ-CLR | ئاوەڵناوی ڕەنگ | 95 | Gerund | Gerund Simple | GRD-S | چاوگی سادە |
| 47 | Adjective | Relational Adjective | ADJ-REL2 | ئاوەڵناوی ڕواڵەت | 96 | Gerund | Gerund Extended | GRD-EXT | چاوگی دارێژراو |
| 48 | Adjective | Participle Adjective | ADJ-PPP | ئاوەڵناوی کراو | 97 | Gerund | Gerund Compound | GRD-COMP | چاوگی لێکدراو |
| 49 | Adjective | Attributive Adjective | ADJ-ATT | ئاوەڵناوی بکەری | | | | | |

The study is also tried to make a standardized POS tagset according to the standardized Universal Dependencies (UD) part of speech. The UD formalism [10] has a POS-tagging standard for most languages. The use of UD allows us to have cross-linguistic consistency and easy communication with existing NLP tasks. The examples of UD formalism are spaCy, Stanza, and UDPipe, which are trained on UD data. We align our tagset with a widely-used POS tagging standard—specifically the UD POS tagset—to ensure cross-linguistic consistency and ease of reference in existing NLP integration. The mapping between proposed POS tagset and the UD is shown in Table II.

It can be noticed that the last POS tag in the mapping is named UNKNOWN (UNK), this is to make guarantee during the POS tagging that if the tokens (or words) in CKL text doesn't belong to any tag of the POS tagset. This is also providing more accurate tagging to ensure that the CKL text might be either has grammar mistake or typos error. These tags help in achieving finer morphological and syntactic granularity which is crucial for processing Kurdish language in an easy and efficient way. This mapping is designed to retain Kurdish texts with our POS tags that are compatible with multilingual treebanks and dependency parsing models to make Kurdish language resources more useful in NLP tasks. The tree structure of the proposed POS tagset is demonstrated in figure 2.

*TABLE II. The mapping between proposed POS tags and the UD equivalent.*

| **Proposed POS Tags** | **Kurdish Label** | **UD Equivalent** |
|---|---|---|
| N-S, N-EXT, N-COMP, N-PROP, N-COM, N-DNP, N-INP, N-SG, N-PL, N-M, N-F, N-GN, N-DUAL, N-CN, N-AB, N-COLL, N-SUB | ناو | NOUN (N) |
| PR-DEM, PR-INT, PR-INDEF, PR-REF, PR-POSS, PR-INDEP, PR-CLIT, PR-QUAN, PR-DEF, PR-S, PR-COMP, PR-EXT, PR-NEG | جێناو | PRON (PRON) |
| ADJ-CLR, ADJ-DESC, ADJ-SUPL, ADJ-COMP, ADJ-QUANT, ADJ-S, ADJ-EXT, ADJ-DIM, ADJ-REL, ADJ-REL2, ADJ-COMPOUND, ADJ- INDEF, ADJ-BASE, ADJ-PPP, ADJ-DEG, ADJ-ATT | ئاوەڵناو | ADJ (Adjective) |
| ADV-S, ADV-COMP, ADV-EXT, ADV-CONJ, ADV-MAN, ADV-QUANT, ADV-NEG, ADV-EMPH, ADV-REP, ADV-CAUS, ADV-LOC, ADV-TIME | ئاوەڵکار | ADV (Adverb) |
| NUM-CARD, NUM-EXT, NUM-COMP, NUM-ORD, NUM-FRAC, NUM-S | ژمارە | NUM (Numeral) |
| V-S, V-EXT, V-COMP, V-PST, V-NPST, V-NEG, V-INCOMP, V-COMPLETE, V-AFF, V-DECL, V-INFORM, V-IMPER, V-PERF, V-IMPERF | کار | VERB (Verb) |
| GRD-D, GRD-W, GRD-Y, GRD-T, GRD-A, GRD-S, GRD-EXT, GRD-COMP | چاوگ | GRD (Gerund) |
| PART-COND, PART-TOP, PART-ADV, PART-LOC, PART-COND, PART-REQ, PART-SURP, PART-EXCL, PART-POL, PART-EMPH, PART-INT | ئامراز | PART (Particles) |
| UNK | UNKNOWN | X (Unknown Category or Stop words) |

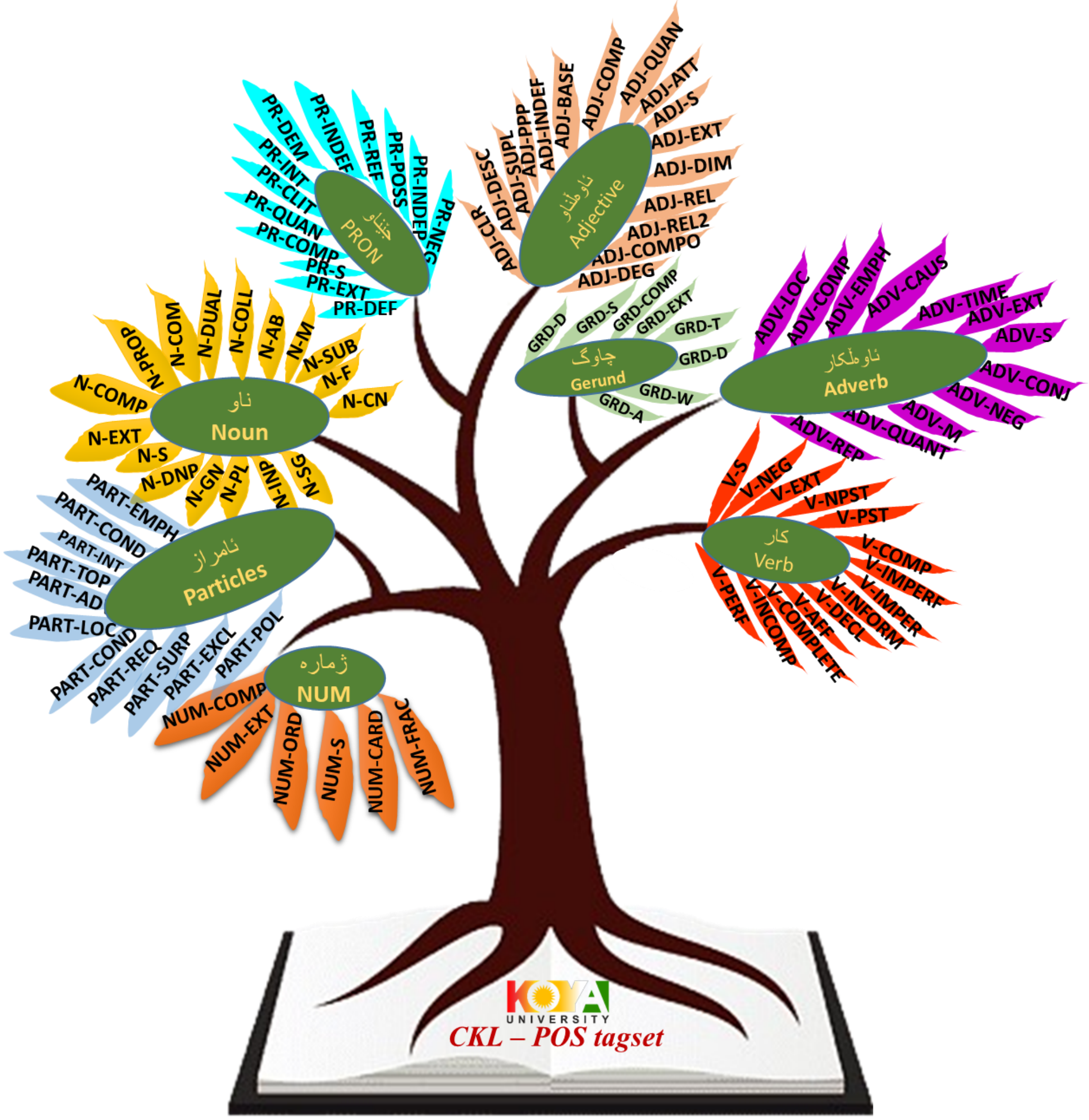

*Figure 2. The tree structure of the proposed CKL - POS tagset.*

To provide a clearer understanding, the following subsections present the details and examples of the word/token structure and the POS tags in CKL text.

### *A. The Word Structure in CKL*

The word structure in CKL text, like other language, could be constructed from different components including: root, prefix, suffix, compound affix, transition word, stem, and base. To this end, figure 3 depicts the scenarios of the word components to show how words/tokens are formed within the CKL text.

1- **Root**: The smallest part of a word, that remains when it is stripped down is called the root; for instance**,** '**بکوژ**' has the root '**کوژ**', which means 'kill', and '**بەهێز**' has the root '**هێز**', which means 'power' [27], [28].
2- **Prefix**: A part that is added at the beginning of a root word and changes its meaning; for example, "**بە**" in the word "**بەهێز**" (***strong***) [27], [28], [29].
3- **Suffix**: A part that is added at the end of a root word and changes its meaning, for example, "**گەر**" in the word "**ئاسنگەر**" (***blacksmith***) [27], [28], [29].
4- **Compound Affixes**: The combination of a prefix and a suffix with a word into a single form that changes the word meaning; such as, '**هەڵگرتنەوە**' (***Lifting***) [27], [28].
5- **Transitions Word**: The result of combining two or more parts of words into one; for instance, '**و**' in '**جل و بەرگ**' (***clothes***) [30].
6- **Stem**: The base form of a word that carries its core meaning, which is simplified into the stem. For example, "**نانەکە**' (***the bread***), where "**ەکە**" means "***the***" and "**نان**" (***bread***) represents the simple stem [27], [28].

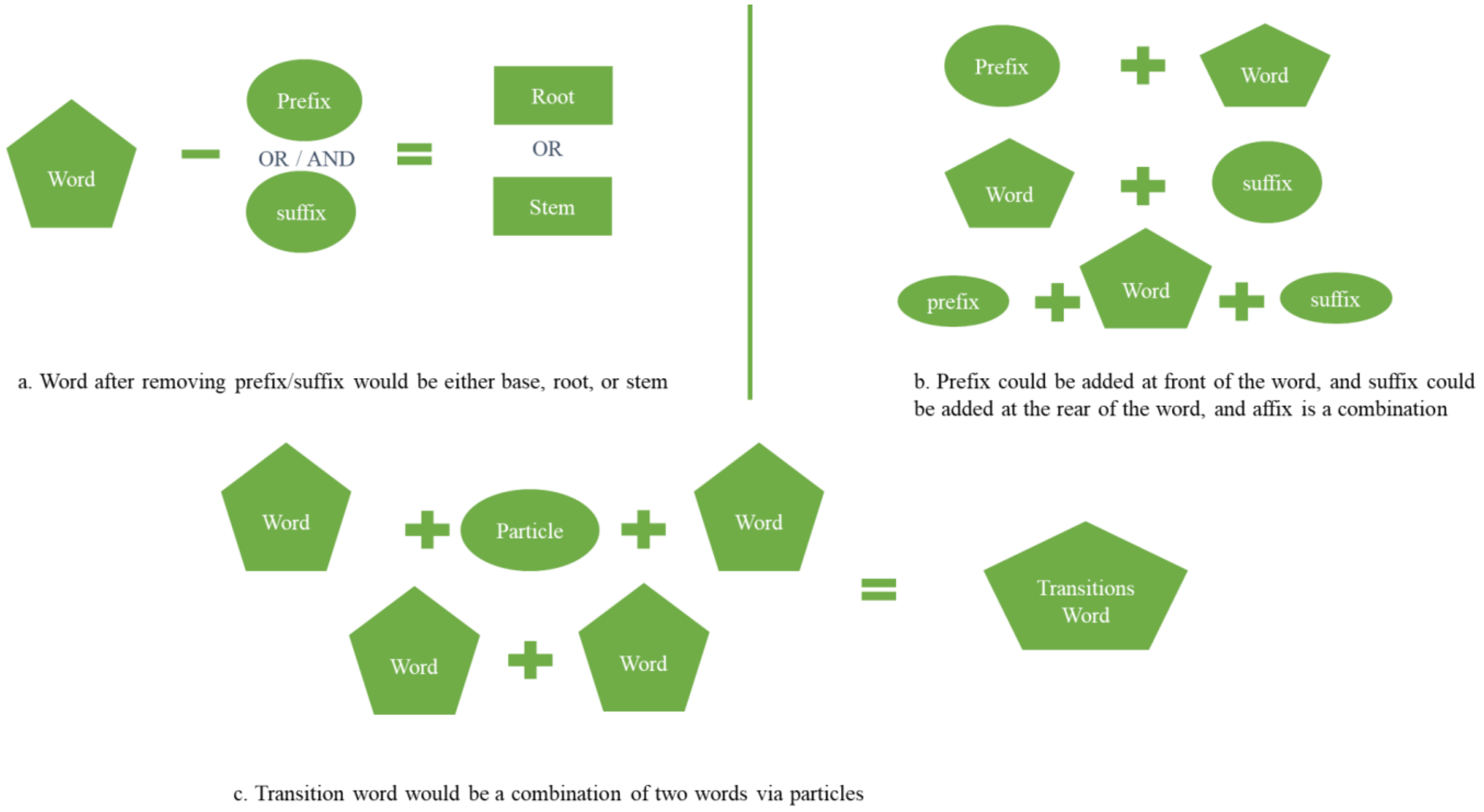

a. Word after removing prefix/suffix would be either base, root, or stem

b. Prefix could be added at front of the word, and suffix could be added at the rear of the word, and affix is a combination

c. Transition word would be a combination of two words via particles

*Figure 3. The word components scenarios in CKL text.*

### *B. The Parts of Speech in CKL*

The parts of speech within the proposed POS tagset for CKL text are categorized into NOUN (N), PRON (PRON), ADJ (Adjective), ADV (Adverb), NUM (Numeral), VERB (Verb), GRD (Gerund), and PART (Particles). The following are explanation of how the categorized part of speech are frequently utilized in CKL text.

**1-** **Noun**: A word used to name a person, thing, or concept, to illustrate (***grape, tree, earth, strength..., etc***.) [22], [25], [27], [29], [30].

- **Simple Noun**: A word that consists of a standalone morpheme, without any prefixes or suffixes, considering: (***flower, city, spring..., etc.***).
- **Derived Noun**: A noun made by adding a prefix, suffix, or both to a simple noun, which changes the meaning or function of the word. For instance, in Kurdish, derived nouns are often formed with the help of suffixes, including: (***question, learner, beauty..., etc***.).
- **Compound Noun**: A noun formed by combining two simple words, with or without a connecting particles, such as: (***kingdom***)***, flowerbed, sunset***).

**A. Nouns Based on Their Meaning** [22], [25], [27], [30]**:**

- **Proper Noun**: A word used to name a specific individual or thing, indicating a particular person or place, such as (***Nasreen, Halgurd, Erbil..., etc.).***
- **Common Noun**: A word used to refer to any person, thing, or concept without distinguishing a specific one, such as (***girl, mother, earth...***, ***etc.).***
- **Collective Noun**: A word that refers to a group or collection of people or things, such as (***people, team, army..., etc.).***

**B. Nouns Based on Number** [22], [25], [27], [30]**:**

- **Singular Noun**: A word that refers to only one individual or thing, for example (***apple, apple tree, worker... etc.***).
- **Plural Noun**: A word used to refer to more than one individual or thing, such as (***women, children, students, etc.***).

**C. Nouns Based on Gender** [22], [25], [27], [30]**:**

- **Masculine Noun**: A word used to refer to a male or masculine role, such as (***Azad, father, rooster..., etc***.).
- **Feminine Noun**: A word used to refer to a female or feminine role, considering (***Namam, mother, chicken... etc***.).
- **(Common Gender Noun):** A word that refers to both male and female, for example (***teacher, person, lamb***).
- **Neuter Noun**: A word used to refer to something without gender, neither male nor female, such as (***bread, floor, door... etc.).***

**D. Nouns Based on Existence** [22], [25], [27], [30]**:**

- **Concrete Noun:** A word used to refer to something that has real, physical existence and can be perceived by the senses, including (**tree, dove, cake, etc.).**

- **Abstract Noun**: A word used to refer to something that does not have physical existence but exists in thought or imagination and cannot be perceived by the senses, such as ***(shame, brotherhood, friendship, etc.)***.

E. **Nouns Based on Recognition** [22], [25], [27], [30]**:**

- **Recognizable Noun**: A word that refers to something that can be clearly identified. When the morpheme *"ەکە"* (***the***) is added at the end of a word, it forms a recognizable noun. For example, (***the flower, the lamp, the book, etc.)***.
- **Indefinite Noun**: A word that refers to something that cannot be clearly identified. These are formed by adding morphemes (***a, an***) *"ێ, ێک, ەک"*. For example, (***an apple, a girl, a door, a room***).

2- **Pronoun: A word used in place of a person's or thing's name, serving as a substitute for the noun in a sentence** [23], [25], [30]**. Types of Personal Pronouns:**

- **Independent Personal Pronoun**: A pronoun used independently in a sentence, without needing any other word to form a meaningful expression. It directly replaces a noun. Examples include: (***I, we, you, they, he, she***).
- **Dependent Personal Pronoun:** A pronoun that is used alongside the name of a person or thing and requires another word to complete its meaning. It cannot stand alone but conveys meaning when attached to another word. Examples include: (me, us, you, them, him, her, etc.).
- **Demonstrative Pronoun**: A pronoun used to refer to specific people or things, including: (***this, that, these, those***).
- **Reflexive Pronoun:** A pronoun used when the subject and object of the sentence are the same. These pronouns show that the action is reflected back onto the subject. Examples include: (myself, ourselves, yourself, themselves)
- **Possessive Pronoun:** A pronoun that indicates ownership or possession or expresses a relationship between something and a person. For example: 'This pen is yours**.**
- **Interrogative Pronoun**: A pronoun is used to ask questions about a person or thing. Examples include: (***who, what, which***).
- **Negative Pronoun**: A pronoun used to indicate the non-existence or absence of a person or thing. Examples include: (***no one, nothing, nobody is arriving***).
- **Quantitative Pronoun**: A pronoun used to refer to quantities or amounts of things. Examples include: ***(many, few, several***) [27].
- **Definite Pronoun**: A pronoun used to refer to specific known persons or things. Examples include: (***everyone, all***) [23], [25], [30].
- **Indefinite Pronoun**: A pronoun used to refer to unspecified or unknown persons or things. Examples include: (***someone, everyone, anyone***) [23], [25], [30].

**Pronouns Based on Composition** [22]**:**

- **Simple Pronoun**: A pronoun formed from a single word that has an independent meaning, such as (***who, which, he, she, etc.).***
- **Derived Pronoun**: A pronoun formed by adding a prefix or suffix to a simple pronoun, such as ***(this one, that one, etc.).***

- **Compound Pronoun**: A pronoun formed by combining two or more words, such as ***(everyone, no one, etc.).***

**3- Particles (Markers):** Particles are used to connect words within a sentence as needed [25], [27], [30].

- **Prefix Particles:** Particles that are attached to the beginning of a word to form the necessary meaning, considering (ی, ە).
- **Linking or Relational Particles**: These are used to form a relationship between a preceding word and a subsequent one, for instance: ***(with, to, for, etc.).***
- **Intervening Particles**: These particles appear between words, for example: (***in... at, with... and..., etc***.).
- **Conjunction Particles**: These are used to connect words or phrases together, namely (***and***).
- **Question Particles**: These particles are used to ask questions, including: (***is, why, etc***.).
- **Conditional Particles**: These particles are used in conditional sentences to indicate conditions and responses, for example: (***if, until, unless, etc.).***
- **Desire Particles**: These particles are used to express desire, hope, or aspiration, considering: (***wish, if only, desire, etc.***) [31].
- **Exclamation Particles**: These words or particles are used to express a strong feeling or emotion, including: (***ah, oh***).
- **Call Particles**: These particles are used to call attention or to emphasize, such as (***hey, yes, oh***).

**4- Adjectives:** Adjectives are words used to describe or modify a noun or pronoun in a sentence, such as (***beautiful, good***), providing more details [25], [27], [29], [30], [32].

- **Relative Adjectives**: These adjectives are formed by adding morphemes to a base word, and they express a relation to something. For instance, "*زێرین*" (***golden***).
- **Indefinite Adjectives**: These adjectives do not directly specify a particular thing but refer to something more general or indefinite. For example: (***other, some***).
- **Qualitative Adjectives**: These adjectives describe the quality or characteristic of something in terms of its nature, appearance, condition, size, color, or other attributes. To illustrate: (***expensive, green, big***, etc.).
- **Quantitative Adjectives**: These adjectives indicate quantity or number. For example: (***one by one, class by class***).
- **Demonstrative Adjectives**: These adjectives show or indicate a specific noun. For instance: (***this, that, these, those***).
- **Interrogative Adjectives**: These adjectives are used to ask questions about the quality or characteristics of something, such as: (***how, which, etc.***).

A. **Adjectives Based on Degree** [25], [27], [30], [32]**:**

- **Positive Adjectives**: These adjectives express a simple or basic quality of a noun or pronoun without any comparison. For example: (***smart, easy...***).
- **Comparative Adjectives**: These adjectives are used to compare two things or people. They are formed by adding the morpheme (***er***) to the adjective. For instance: (***smarter, easier***).

- **Superlative Adjectives**: These adjectives express the highest degree of quality, used to compare one thing to a group of things. They are formed by adding the morpheme (***est***) to the adjective. such as: (***smartest, easiest***).

B. **Adjectives Based on Composition** [25], [27], [29], [30], [32]**:**

- **Simple Adjectives**: These adjectives are formed from a single word expressing a quality or characteristic independently. For example: (***good, happy***, etc).
- **Derived Adjectives**: These adjectives are formed from a simple word by adding a prefix or suffix. They include a prefix or suffix that modifies the original meaning. Such as: (***sad, unconscious, strong, etc.***).
- **Compound Adjectives**: These adjectives are formed by combining two or more words to describe something. For instance: (***glass-like, clear-sighted, etc***.).

**5-** **Adverbs:** Adverbs are words used to describe or modify the action, state, or condition of a verb, adjective, or another adverb. They often express time, place, manner, or degree and are used to provide more detail about the action or situation [25], [27], [30], [33].
**Types of Adverbs:**

- **Adverbs of Time**: These adverbs indicate the time of an action or event. For example: (***now, today, later,*** etc.).
- **Adverbs of Place**: These adverbs specify the location of an action or event. For instance: (***near, above, below, etc.).***
- **Adverbs of Manner**: These adverbs describe how an action is performed. To illustrate: (***badly, correctly – The child fell badly***).
- **Adverbs of Quantity**: These adverbs indicate the degree or extent of an action or event. This includes (***a lot, a little – The child came very quickly, etc.***).
- **Adverbs of Negation**: These adverbs indicate the negation or absence of an action. For example: ***(never, in no way, etc.).***
- **Adverbs of Effort**: These adverbs describe the effort or exertion put into achieving a result. For example: (***I always try***).
- **Adverbs of Repetition**: These adverbs indicate that an action is repeated. For instance: (***again, once more***).
- **Adverbs of Cause and Purpose**: These adverbs express the cause or purpose behind an action. Take, for example: (***because of, due to, etc.).***
- **Adverbs of Arrangement**: These adverbs describe the manner or organization in which an action is carried out. Such as: (***the students came one by one***).

**Adverbs Based on Composition** [25], [27], [30], [33]**:**

- **Simple Adverb**: These adverbs are formed from a single word that independently expresses an action's manner, time, or place. For example: (***quickly, now, below***, etc.).
- **Derived Adverb**: These adverbs are formed from a simple word by adding a prefix or suffix. They describe the action in relation to another element. For instance: (***carefully, courageously, like a lion***).

- **Compound Adverb**: These adverbs are formed by combining two or more words to describe an action in greater detail, including (***step by step, continuously, downward***, etc.).

6- **Number:** A number is a term used to indicate quantity, ratio, and the size of objects within a sentence [25], [27], [30], [33].

A. **Numbers Based on Meaning:**

- **Cardinal Number**: These numbers express the exact amount or count of something. For example: (***I read two books***).
- **Ordinal Number**: These numbers are used to show the position or order of something. For instance: (***first, third***, etc).
- **Fractional Number**: These numbers express a portion or fraction of something. Take, for example: (***half, one-quarter...***)

**B. Number Based on Composition** [27], [30], [33]:

- **Simple Number**: These numbers consist of a single, standalone numeral. For example: (***one, ten, thirty***, etc.).
- **Derived Number**: These numbers are formed by adding a prefix or suffix to a simple number to create a new value. For instance: (eighty, fifty, etc.)
- **Compound Number**: These numbers are formed by combining two or more numbers to indicate a specific quantity or amount, namely (***fourteen, two thousand***, etc.).

7- **Gerund (Chawg):** A verb base is the core form of a verb that indicates an action but does not express tense. It is the form that is used as the foundation for conjugating the verb into different tenses, moods, or persons. It doesn't directly indicate time and is not tied to a specific subject [25], [27], [30]. In Kurdish, there are five types of verb bases based on the prefix used [25], [27], [30], [33]:

- **Gerund Dali**: The verb base with the prefix "**د**" (***d***) before it. For example: (***خوێندن - to read***).
- **Gerund Wawi**: The verb base with the prefix "**و**" (***w***) before it. For instance: (***چوون - to go***).
- **Gerund Yai**: The verb base with the prefix "**ی**" (***y***) before it. Such as: (***کڕین - to buy***).
- **Gerund Tai**: The verb base with the prefix "**ت**" (***t***) before it. To illustrate: (***هاتن - to come***).
- **Gerund Alfi**: The verb base with the prefix "**ا**" (***a***) before it. For example: (***شکان - to break***).

8- **Verb:** A verb is a word that indicates an action and the time in which it occurs. It is used to refer to a person performing an action. For example: (***eating, coming...)*** *[24], [25], [27], [30]***.**

A. **Verb Based on Composition** [25], [27], [30], [33]**:** According to new trends in linguistics, a verb is analyzed based on its morphemes. In its basic form, a verb may not have a simple action in the sentence since it usually includes a tense marker. However, if we analyze it as a word, several types exist:

- **Simple Verb**: These verbs are formed from a single word with an independent meaning. For example: (***to carry, to hold, etc.).***
- **Derived Verb**: These verbs are formed by adding a prefix or suffix to a simple verb, such as (***to intend, to close, etc.).***
- **Compound Verb**: These verbs are formed by combining a simple verb with another verb or a root. For instance: (***Forget it, to cheer up, etc.).***

B. **Verb Based on Time** *[24], [25], [27], [30]***:**

- **Past Action**: This refers to an action that has already occurred and is finished, happening before the present moment. For example, (***ate, washed, chose, wrote, and so on).***
- **Future or Ongoing Action**: This refers to an action that is expected to happen after the present moment or is happening in the future or is ongoing. For instance, (***will sleep, will choose).***

C. **Verb Based on Effect** *[24], [25], [27], [30]***:**

- **Effecting Action**: This refers to an action where the effect of the action directly impacts an object, typically with a direct and immediate result. For example, (***The child ate the apple***).
- **Non-Effecting Action**: This refers to an action where the effect does not immediately impact another action, or the effect is not directly observable in the same sentence. For instance, (***The child slept***).

D. **Verb Based on Completion** *[24], [25], [27], [30]***:**

- **Complete Action**: This refers to an action where the time and event are fully completed, and the action directly conveys a sense of completion with no further need for additional action. The verb is fully self-contained, for example, ***(I wrote my story***).
- **Incomplete Action**: This refers to an action that is not fully completed on its own, often requiring the addition of another verb to complete its meaning. The action is not fully carried out until the additional word or action is added, for instance: (**must go, need help, should eat**)**.**

E. **Verb Based on Potentiality** [24], [25], [30]**:**

**Potential Action**: This refers to an action that is possible or feasible, and the ability to perform the action is inherent in the verb itself. It indicates that the action can be carried out. For example: ***(can write, can go, ...).***

- **Impossibility or Negative Potential Action**: This refers to an action that is not possible or cannot happen, or when the action is negated. Sometimes, it involves a modification that prevents the action from occurring, including: (***cannot write, cannot go, unable to***).

F. **Verb Based on Style and Mode** [24], [25], [30]**:**

- **Direct Mode of Action**: This refers to a direct style where the action is clearly expressed and affects a person or thing. It shows the completion of an action or event, for instance, (I sent my book to my friend).
- **Indirect Mode of Action**: This refers to a style where the action is not directly stated but involves a subjective or speculative aspect, often involving intention, hope, or desire. For example: (***I wish, I hope, I intend, etc.).***
- **Command Mode**: This refers to the mode in which one person ***(the speaker)*** gives an order or request to another person (the listener) for an action to be performed. The speaker is giving a command to complete an action, such as: ***(Read, sit down, write).***

### *C. Kurdish NLP*

The NLP is a branch of artificial intelligence (AI) that enables computers to comprehend, analyze, and produce human language. It combines insights from linguistics, computer science, and machine learning to manipulate both written and spoken data in ways that approximate how humans understand language. NLP is basically the effort to make human language computable and useful in real-world applications [19]. To do end, there are several NLP tasks which are broadly categorized into two categories: **syntactic processing** and **semantic processing**. The syntactic processing deals with the structure of word sequences, including part-of-speech tagging, parsing, and tokenization. In contrast, the semantic processing involves the meaning of words, such as named entity recognition, sentiment analysis, and question answering. For these tasks, they need large, annotated corpora to train models that capture linguistic patterns and structures [20]. The NLP is still in its early stages for low-resource languages such as Kurdish language (including all its dialects). Difficulties like the lack of standardized orthography between different dialects and limited annotated data via an accurate and standardized POS tagset. This is for additional processing, which has made it hard to computationally process the Kurdish text. Thus, constructing such an accurate and standardized POS tagset is a very initial step to help Kurdish NLP tasks.

Furthermore, a comprehensive and accurate POS tagset will be a benchmark for most of the current Kurdish NLP task including search engines, intelligent assistants, automatic translation, and more. The following are the most popular and influential tasks of NLP [19]:

**1. Machine Translation**: Machine Translation (MT) is the process of automatically translating text or speech from one language into another. Google Translate, DeepL and Microsoft Translator are systems that translate documents, web pages or conversations in real time, using statistical or neural models. The MT is important for cross-lingual communication, multilingual content retrieval, and global collaboration.

**2. Named Entity Recognition (NER):** This can identify information like names, organizations, dates, and locations contained within text.

**3. Syntactic Parsing:** Identifies grammatical constructs like noun and verb phrases or who did what to whom.

**4. Word Sense Disambiguation (WSD):** Given a word, you can choose the right meaning based on its context.

**5. Text-to-Speech (TTS):** TTS systems convert written text into natural speech, basically enabling computers to have the ability to "speak" text. This is useful in screen readers, tools for students who are learning how to read, audiobooks, and applications for users with impaired eyesight.

**6. Sentiment Analysis:** Analyzing sentiment to identify the underlying emotional tone or opinion present in a piece of text [21]. Sentiment analysis is commonly used in social media monitoring, customer feedback analysis, brand reputation management, and mining of product reviews to identify positive, negative, or neutral sentiments.

**7. Text Summarization:** Automated text summarization algorithms produce a short schematic of a document, preserving its main aspects [21]. There are two types of text summarization 1) Extractive summarization, which pulls important sentences. 2) Abstractive summarization expresses the content with new sentences. This is helpful in news aggregation, legal documents, academic research, and healthcare reports.

**8. Document Classification:** This is the task of mapping text to one or more of the predefined categories. Common examples include spam detection, news or blog topic classification, and legal or medical document categorization.

The aforementioned NLP tasks could be implemented for the Kurdish language (including all the dialects), if an accurate or comprehensive POS tagset is utilized. However, according to the state-of-the-art, there is no a standardized or comprehensive Kurdish POS tagset, yet. Therefore, this study proposed a new and accurate POS tagset for Central Kurdish Language, considering many of the requirements of NLP tasks within the tagset. To understand the process of the POS tagging for CKL tokens/words in a sentence, figure 4 illustrates an example of how the tokens are tagged via our proposed POS tagset. This is followed by utilizing the tagged tokens for the Kurdish NLP tasks.

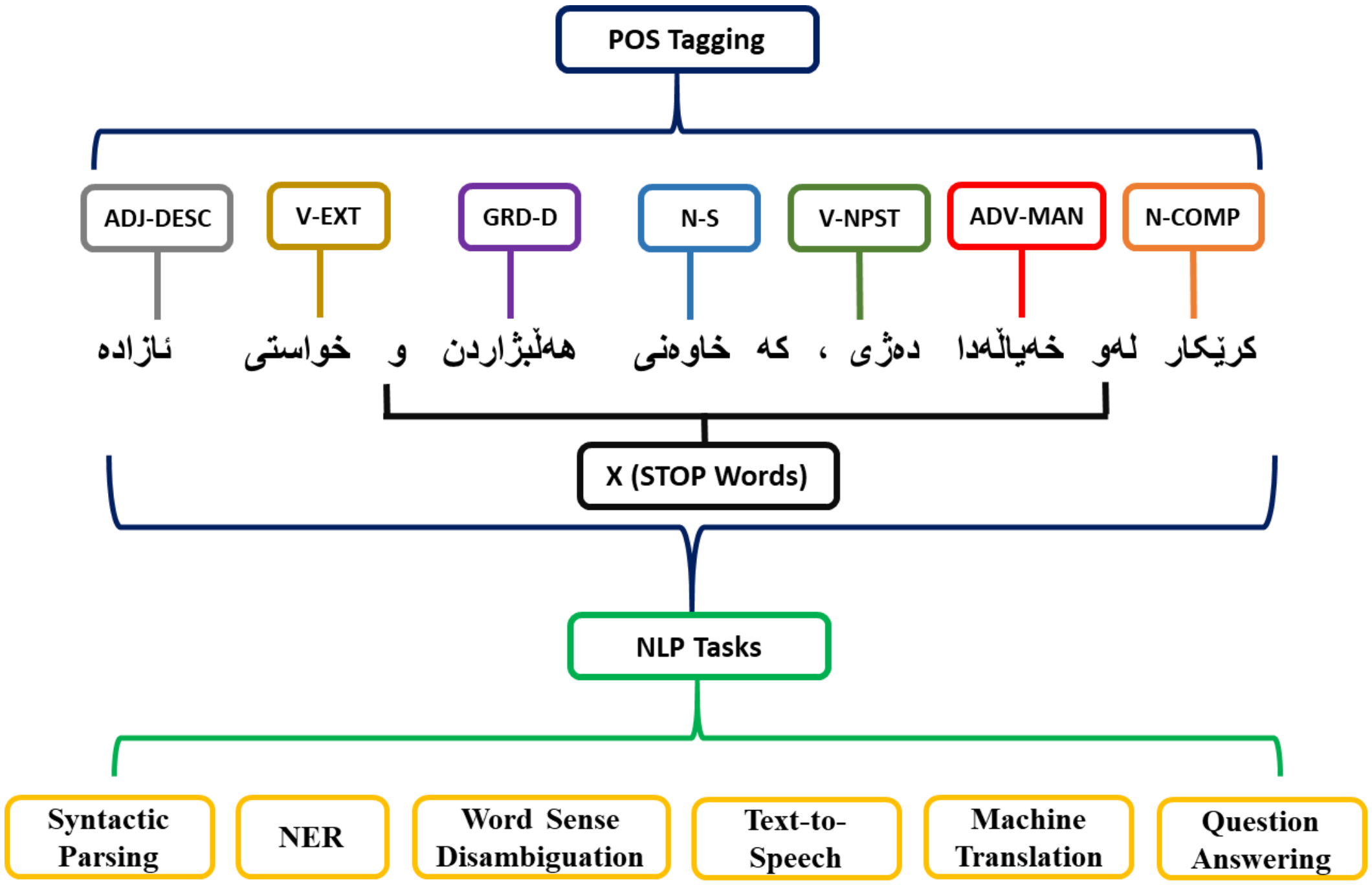

*Figure 4. An example of POS tagging for CKL sentence as an input for Kurdish NLP.*

## 5- Conclusions and Future Steps

Creating a POS tagset for the CKL is a crucial advancement for NLP tasks in this underrepresented language. A significant challenge faced in this study was the limited availability of resources to demonstrate a comprehensive part of speech tagset for CKL. This scarcity has long impeded the development of robust NLP tasks for CKL, restricting the ability to train high-quality language models for tasks such as part-of-speech tagging, syntactic parsing, and machine translation.

This study addresses these challenges by developing a comprehensive POS tagset tailored to the richness of CKL language, ensuring accuracy and adaptability for various NLP tasks. The tagset reflects the language's specific syntactic and morphological features, such as flexible word order and rich inflectional morphology. Its comprehensiveness allows for the disambiguation of similar syntactic structures, an essential requirement for high-performing NLP tasks.

Given the constraints of large annotated corpora, in the near future, our professional team (from the scientific departments of Koya University) has planned to utilize a corpus constructed from academic documents. This is because the academic texts were selected for their relative consistency in grammar, spelling, and linguistic conventions, making them a valuable resource for proposed POS tagging and followed by training NLP models. Additionally, the higher editorial quality of academic documents provides more predictable sentence structures and fewer errors compared to informal texts, making them ideal for POS annotation.

However, the use of academic texts also presents challenges. The complexity of academic writing and its formal vocabulary may not fully represent the linguistic diversity found in other domains, such as news, social media, or spoken language. Therefore, while the constructed corpus is a valuable resource for POS tagging, it may not capture the full range of linguistic variation present in everyday Kurdish.

Annotating the Central Kurdish Corpus with the new POS tagset will yield invaluable insights into the language's features and their applicability to NLP tasks. Thereafter, the tagged corpus can serve as a foundation for developing more sophisticated NLP models, such as part-of-speech taggers, named entity recognizers, and syntactic parsers, significantly enhancing computational resources for the Kurdish language. Furthermore, the development of this tagset and future annotated corpus is a critical step toward broader research on Kurdish language processing, providing a framework for standardization and consistency in future NLP applications.

Looking ahead, efforts should also focus on expanding the corpus by incorporating data from diverse domains and addressing linguistic phenomena that may not be adequately represented in academic texts. Collaboration with native speakers and linguistic experts will be essential to refine the tagset and ensure its applicability across different Kurdish dialects. The successful completion of this research establishes a solid foundation for advancing Kurdish NLP and paves the way for further exploration of language processing in low-resource languages.

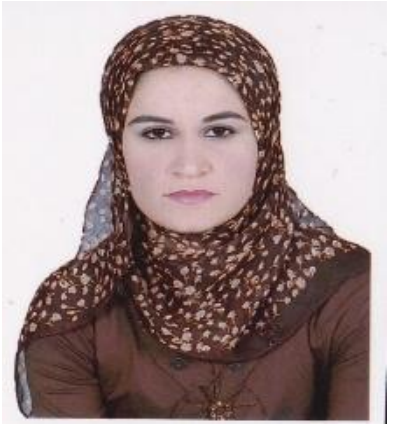

**Shadan Sh. Sabr** is a Lecturer teaching in the Department of Kurdish language, Faculty of Education at Koya University. She started working at Koya university first time in 2009 as an Asst. Researcher after she gained BA in Kurdish Language at Koya University in 2008. She gained MA in 2013 in Kurdish Language at Salahaddin University/Erbil under the title of (Code Switching in Conversation, Erbil city as a sample). also dktora at koya university in 2023 that title of (The Principles of the Term Formation in Kurdish Language Legal Term as an Instance). She currently serves as an Assistant Professor at the same University.

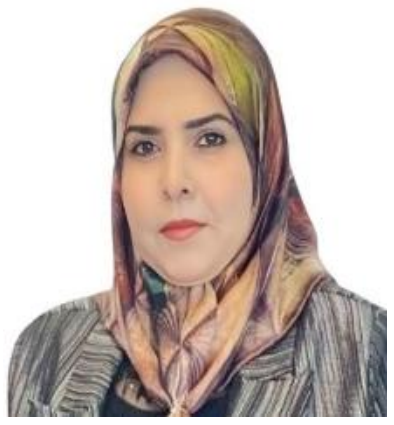

**Nazira S. Mustafa** is a lecturer in the Department of Kurdish Language, Faculty of Education, Koya University. She started his career as a lecturer at Koya University in 2012. She received his bachelor's degree in the Kurdish Language from Salahaddin University in 1995, and then obtained her master's degree in the Kurdish Language from Koya University In 2008., She received his PhD in Philosophy of Kurdish Language from Koya University in 2018 entitled "Pragmatics and Stylistics" She currently serves as a lecturer at the same university.

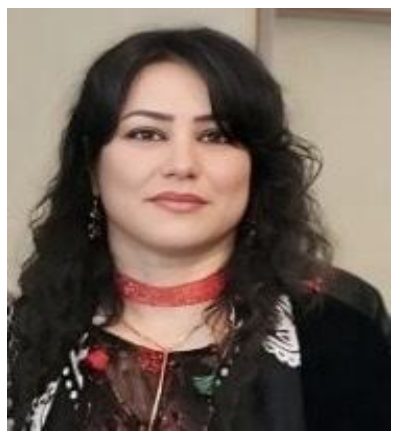

**Talar. S. Omar** is a Head of Department of Kurdish language, Faculty of Education at Koya University. She started working at Koya university first time in 2009 as an Asst. Researcher after she gained BA in Kurdish Language at Koya University in 2009. She gained MA in 2013 in Kurdish Language at Salahaddin University/Erbil under the title of (The Language of Audiovisual Advertisements- an Analytic Study). also dktora at University of Raparin in 2022 that title of (Discourse Analysis of health from the perspective of Cognitive Theory). She currently serves as an Assistant Professor at the same University.

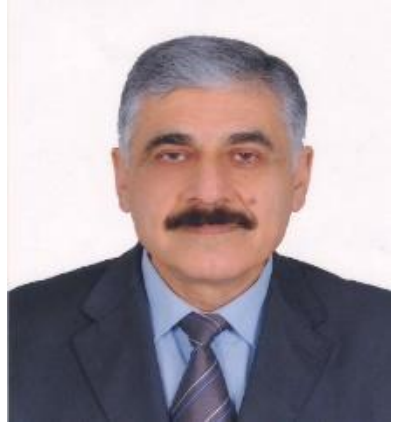

**Salah Hwayyiz Rasool** is a Lecturer teaching in the Department of Kurdish Language, Faculty of Education at Koya University. He gained B.A. in Kurdish Language & Literature at University of Sulaimani in 2004. He started working at Koya University for the first time at 2007, after he gained M.A. in Kurdish Language at Koya University in 2006, under the title “The Principle of Economy in Kurdish Language, From Perspective of Government and Binding Theory and Minimalist Program”. He gained Ph.D. in Kurdish Language at Koya University in 2013, under the title “The Domination of Language over Kurd’s Personality – Proverbs as Example”. He currently serves as an Assistant Professor at the Koya University.

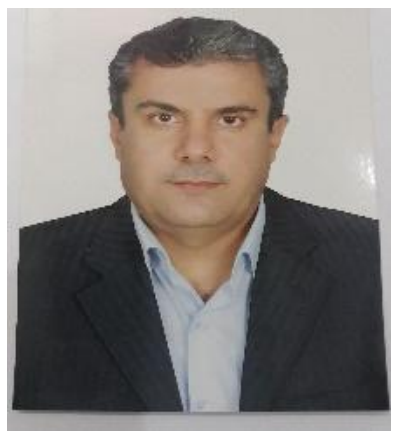

**Nawzad Anwer Omer** is a Lecturer teaching in the Department of Kurdish Language, Faculty of Education at Koya University. He gained B.A. in Kurdish Language & Literature at University of Sulaimani in 2004. He started working at Koya University for the first time at 2010, after he gained M.A. in Kurdish Language at Koya University in 2008, under the title “The Causative in Kurdish Language”. He gained Ph.D. in Kurdish Language at University of Sulaimani 2014, under the title “Some Aspects of socio-pragmatic in Kurdish Language”. He currently serves as a Professor at the Koya University.

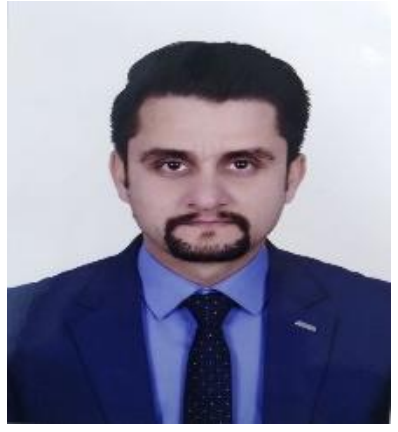

**Darya S. Hamad** is a Lecturer teaching in the Department of Kurdish language, Faculty of Education at Koya University. He started working at Koya university first time in 2011 as an Asst. Researcher after he gained BA in Kurdish Language at Koya University in 2011. He gained MA in 2016 in Kurdish Language at University of Raparin/Rania under the title of (Phonetic Issues of Kurdish Language between Orthography and Phonological Rules). Also PhD at Koya University/Koya in 2023 that title of (Comparison of Morphological Processes between Kurdish and Persian from the Perspective of Morpheme-based Morphology Theory). He currently serves as an Assistant Professor at the same University.

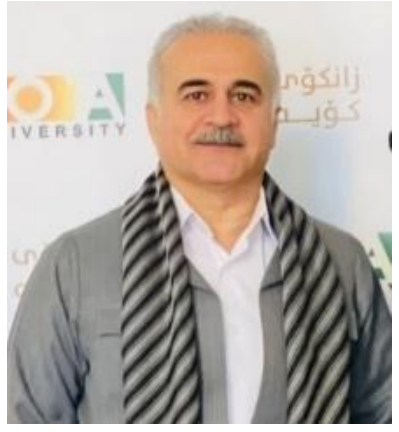

**Hemin Abdulhameed Shams** is a Lecturer teaching in the Department of Kurdish Language, Faculty of Education at Koya University. He gained B.A. in Kurdish Language & Literature at University of Sulaimani in 2004. He started working at Koya University for the first time at 2007, after he gained M.A. in Kurdish Language at Koya University in 2006, under the title “Style and Expression on Social Occasions”. He gained Ph.D. in Kurdish Language at Koya University in 2013, under the title “The Domination of Language over Kurd’s Personality – Proverbs as Example”. He currently serves as an Assistant Professor at the Koya University.

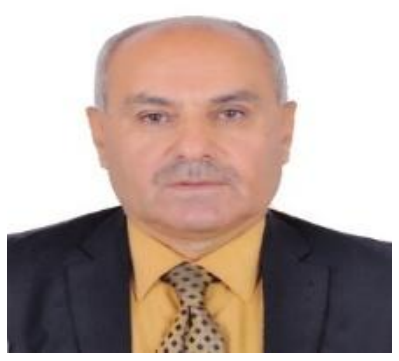

**Omar Mahmood Kareem** is a lecturer teaching in the Department of Kurdish Language, Faculty of Education at Koya University. He gained B.A in Kurdish Language and Literature at Salahaddin University in 1988. He gained M.A in Kurdish Language at Bagdad University in 2003, under the title (The transitive and intransitive verbs in Kurdish). He started working at Koya University for the first time in 2004. He gained Ph.D. in Kurdish Language at Koya University at 2009, under the title (The Pragmatic and Semantic features of Presupposition). He currently serves as an Professor at Koya University

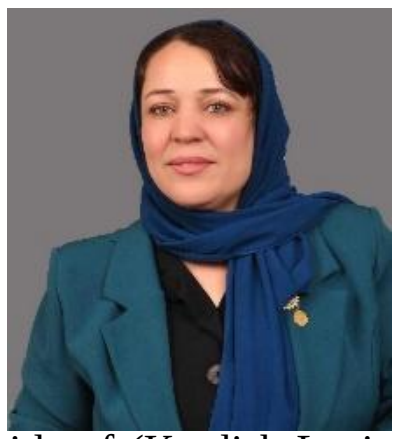

**Rozhan N. Abdullah** is a lecturer teaching in the Department of Kurdish Language, Faculty of Education, Koya University. She started Working at Koya University first time in 2006. As an Asst. Researcher after she gained BA in Kurdish Language at Salahaddin University in 1999. She gained MA in 2004 in Kurdish Language at Koya University under the title of (Kurdish Lexicon and Terminology). Also dktora at Koya University in 2014 that title of “Phenomenon of metaphor and its role in enriching language”. She Currently serves as an Assistant Professor at the same University.

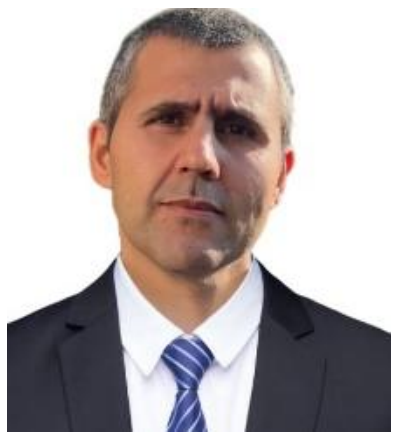

**Khabat Atar Abdullah** is an Instructor and Researcher with a Master of Science in Computer Engineering (Software) from Islamic Azad University, Science and Research Branch, Tehran, Iran (2021). He also holds a Bachelor of Science in Computer and Static from the University of Sulaimani, Iraq (2009). Since 2024, he has been serving as an Assistant Lecturer at Koya University, Faculty of Science and Health in the Department of Computer Science. He currently holds the position of Assistant Lecturer at the same university.

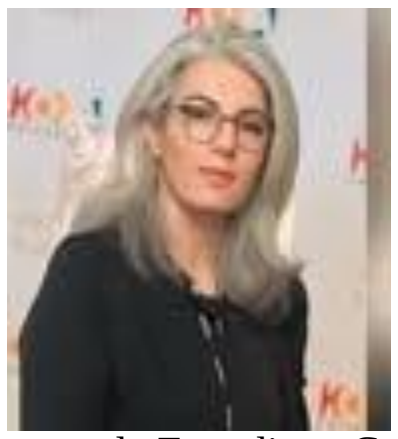

**Mahabad Azad Mohammed** is a lecturer teaching in the Department of Kurdish Language, Faculty of Education, Koya University. She is the director of postgraduate directorate of Koya University. He is also working as a researcher after she gained PhD in Philosophy of Literature and Kurdish Literary Criticism. She was a member of several Founding Committee and permanent member of the Department of Great Mullah Research Center at Koya University. She has participated, organized and prepared several conferences, scientific meetings and workshops in the Great Mullah center. Last but not leas she is the director of Zainab Khan Institute for Research and Translation.

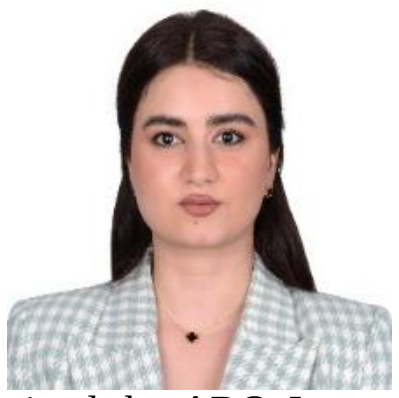

**Haneen Al-Raghefy** is the Head of the Secretarial Office at the ARO Journal at Koya University. She earned her B.Sc. in Software Engineering from Koya University, Erbil, Iraq, in 2016, graduating top of her class. She began her academic career in 2017 as a Laboratory Assistant at Koya University and later joined the ARO Journal, where she continues to contribute to the journal's editorial operations. She is currently pursuing her M.Sc. in Software Engineering, with a research focus on Artificial Intelligence and Natural Language Processing, particularly in the development of resources for low-resource languages. In addition to her academic and administrative roles, she has recently completed two prestigious leadership programs in the United States, reflecting her commitment to academic excellence, cross-cultural collaboration, and professional development.

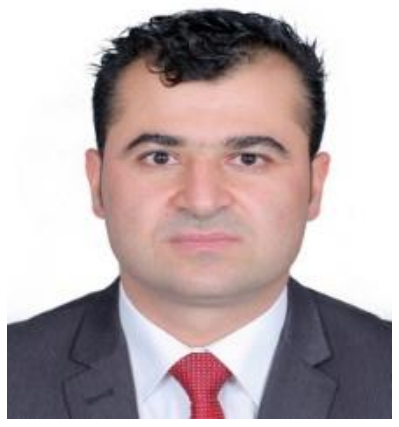

**Safar M. Asaad** is an Instructor and Researcher with a Ph.D. in Information Systems Engineering. He obtained his M.Sc. Eng. in Advanced Software Engineering from the Faculty of Engineering at the University of Sheffield, UK, in 2012. He earned his B.Sc. in Software Engineering from Koya University, Erbil, Iraq, 2008 and completed his Ph.D. at Erbil Polytechnic University, Erbil, Kurdistan Region of Iraq, in 2021. In 2011, he was awarded a scholarship to pursue his studies in the UK through the Human Capacity Development Program (HCDP) by the Ministry of Higher Education and Scientific Research, Kurdistan Regional Government. He completed his postgraduate program in Software Engineering in 2012. Since 2013, he has been affiliated with Koya University, initially as an Assistant Lecturer in the Department of Software Engineering. He currently serves as an Assistant Professor at the same University.

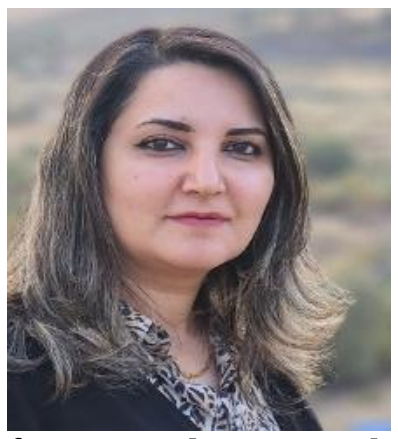

**Sara Jamal Mohammed** is an Instructor and Researcher with a MA in English language and linguistics from the Faculty of Humanities and Art at the Leicester University. She obtained her BA. in English Language from English Department at the University of Sulaimaniyah. She works as assistant researchers after graduation form college for several years. She participated in numbers of international conferences from different local and international universities inside and outside the country. She qualified with training course form Arkansas University/ USA, and works as teaching staff for several years at BELC at University of Koya. She is a PhD student in Applied linguistics at faculty of humanities and social science at the University of Koya and serves as lecturer at Faculty of Education/ Department of English Language at the same university.

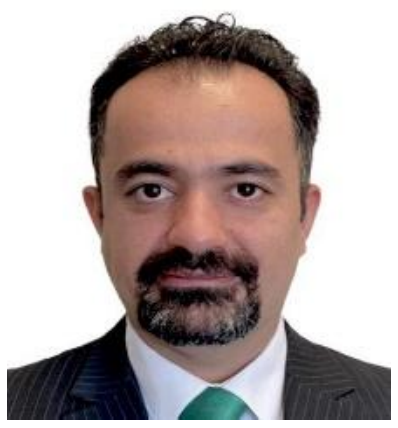

**Twana Saeed Ali** is a lecturer and Researcher with M.Sc. in Management Information System (MIS). He obtained his B.Sc. in Computer Engineering at Koya University, in 2011. He earned his M.Sc. from Cyprus International University (CIU), Nicosia, Cyprus, 2019. He was awarded Cisco Networking Academy and Participating on international Cybersecurity course at Genoa, Italy at 2024. His professional on Data analysis, Management information, IOT, biometric system and e-commerce. Since 2013, he has been affiliated with University of Sulaimani, initially, current role at Sulaimani University is Director of Information Technology. He currently serves as a lecturer at the same University.

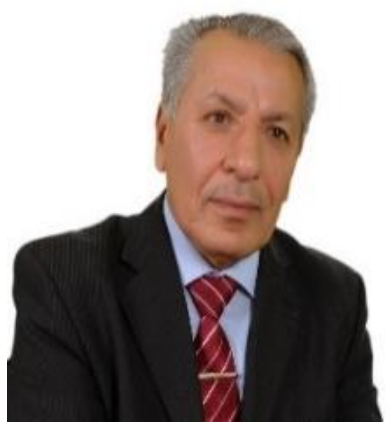

**Fazil Shawrow**, was graduated at Baghdad University in 1975, and received AB Degree in English Language and Literature at College of Arts - English Department. He also got Diploma Degree in Computer Programming at Rathmines College in Dublin (2009). He worked as English Language teacher English for 35 years in (Iraq, Libya and Republic of Ireland). Now he is retired and lives in Dublin. Shawrow has 23 published books in the fields of poetry, national culture, Kurdish language, literary criticism, education, environment and memoir writing. He is active in presenting conferences and seminars. He is now working with Koya University preparing Koya Encyclopedia. He is a member of the Kurdish Writers Union and the editor-in-chief of three literary, cultural and educational magazines.In addition to writing, he is passionate about painting, sculpture and other works of art. He has opened eight exhibitions of miniatures in Kurdistan and Iraq. He has been awarded several prizes and honors. In 2004, The Dublin City Council offered him The Dublin Award which is only given to three active Dubliners annually.

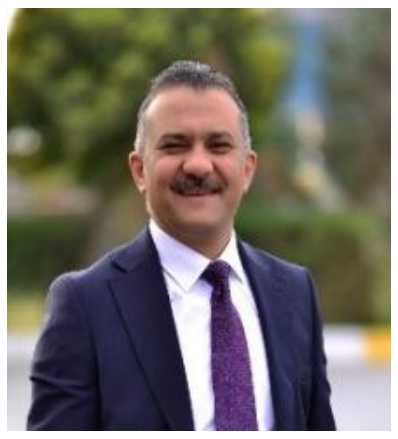

**Halgurd S. Maghdid** received the B.Sc. degree in Software Engineering from Salahaddin University, Erbil-Iraq (2004). And, he received an M.Sc. degree in Computer Science from the Koya University in 2006, Koya-Erbil-Iraq, where continue as an assistant professor till now. In 2016, he got a Ph.D. in Applied Computing at the University of Buckingham, UK. His research focuses on hybrid GNSS with other wireless/sensor technologies including WiFi, Bluetooth, and inertial sensors to offer seamless outdoors-indoors Smartphone localization. He is also working on solving real life problem by using artificial intelligence, deep learning, and machine learning algorithms.